\documentclass[journal]{IEEEtran}
\usepackage[T1]{fontenc}

\usepackage[utf8]{inputenc}
\usepackage{amsmath,amssymb,bm,booktabs,graphicx,microtype,array}
\usepackage{algorithmic,cite,dblfloatfix}
\usepackage{hyperref}
\hypersetup{hidelinks}
\graphicspath{{figures/}}
\newcommand{\R}{\mathbf R}

\newcommand{\fig}[3]{\begin{figure}[!t]\centering\includegraphics[width=\columnwidth]{#1}\caption{#2}\label{#3}\end{figure}}
\title{Two stage GNSS outlier detection for graph optimization based GNSS-RTK/INS/odometer fusion}
\author{Baoshan Song, Penggao Yan, Xiao Xia, Yihan Zhong, Yuan Li, Weisong Wen, and Li-Ta Hsu\thanks{All authors are with the Department of Aeronautical and Aviation Engineering, The Hong Kong Polytechnic University, Hong Kong SAR, China.}\thanks{Corresponding author: Li-Ta Hsu (e-mail: lt.hsu@polyu.edu.hk).}}

\newcommand{\pI}[1]{\mathbf p_{\mathrm I,#1}^{w}}
\newcommand{\pG}[1]{\mathbf p_{\mathrm G,#1}^{e}}
\newcommand{\Rb}[1]{\mathbf R_{b,#1}^{w}}

\newcommand{\bg}[1]{\mathbf b_{g,#1}^{b}}
\newcommand{\ba}[1]{\mathbf b_{a,#1}^{b}}
\newcommand{\thetaB}[1]{\boldsymbol\theta_{#1}^{b}}

\newcommand{\resZeroRTKThreeD}{7.92}

\newcommand{\resZeroRTKSThreeD}{0.30}

\newcommand{\resZeroRISThreeD}{0.28}

\newcommand{\resZeroRIDThreeD}{0.28}

\newcommand{\resZeroRIOSThreeD}{0.21}

\newcommand{\resZeroRIODThreeD}{0.21}

\newcommand{\resZeroZairGain}{58.56}

\newcommand{\resThirtyRTKSThreeD}{1.94}

\newcommand{\resThirtyZairThreeD}{1.99}

\newcommand{\resThirtyRIThreeD}{2.15}

\newcommand{\resThirtyRIDThreeD}{0.38}

\newcommand{\resThirtyRIODThreeD}{0.23}

\newcommand{\resThirtySunGain}{24.84}

\newcommand{\resSixtyRIDThreeD}{0.40}

\newcommand{\resSixtySunThreeD}{1.34}

\newcommand{\resSixtyRIOUp}{1.24}
\newcommand{\resSixtyRIOThreeD}{1.27}

\newcommand{\resSixtyRIOSThreeD}{0.33}

\newcommand{\resSixtyRIODUp}{0.17}
\newcommand{\resSixtyRIODThreeD}{0.27}
\newcommand{\resSixtyRIOGain}{78.51}
\newcommand{\resSixtySunGain}{70.42}

\newcommand{\resGugongRIDThreeD}{0.96}

\newcommand{\resGugongSunThreeD}{0.95}
\newcommand{\resGugongRIOH}{0.18}
\newcommand{\resGugongRIOUp}{0.76}
\newcommand{\resGugongRIOThreeD}{0.78}

\newcommand{\resGugongRIODH}{0.16}
\newcommand{\resGugongRIODUp}{0.62}
\newcommand{\resGugongRIODThreeD}{0.64}

\newcommand{\resMediumRIThreeD}{1.75}

\newcommand{\resMediumRIDThreeD}{1.70}

\newcommand{\resDeepRIUp}{3.61}
\newcommand{\resDeepRIThreeD}{3.90}

\newcommand{\resDeepRIDUp}{2.08}
\newcommand{\resDeepRIDThreeD}{2.37}

\newcommand{\resDeepSunGain}{39.11}
\newcommand{\resTimeRIDMean}{32.03}
\newcommand{\resTimeRIDPninetyfive}{50.80}

\newcommand{\resTimeRIODMean}{79.36}
\newcommand{\resTimeRIODPninetyfive}{126.47}

\newcommand{\resGugongRIOGain}{18.05}

\begin{document}
\maketitle
\begin{abstract}
GNSS pseudorange outliers severely disturb carrier-phase ambiguity resolution and positioning, particularly during signal blockages and recovery. Temporal Doppler consistency provides a reliable short-term range-change reference, but requires valid measurements at both endpoints and becomes unavailable upon signal re-acquisition. This paper presents a two-stage pseudorange screening framework that complements Doppler consistency with joint IMU-odometer motion prediction. The second stage compares predicted and measured temporal changes of double-differenced pseudoranges against an accepted anchor epoch without using current GNSS observations in the motion reference. A first-order error model separates anchor, relative-motion, and measurement uncertainties to propagate factor-graph covariance into the chi-square detection gate, and both stages adjust code weights through a closed-form minimum-inflation rule. Under continuous tracking, Doppler screening reduces RTK 3D RMSE by 96.18\% and outperforms the a-contrario baseline by 58.56\%. When Doppler checks are unavailable, joint IMU-odometer prediction detects a 10-m outlier change in 99.58\% of trials after 30 s and 73.72\% after 60 s, reducing RIO 3D RMSE by up to 78.51\% after outages while RI screening outperforms an inertial-aided baseline by 70.42\% in simulation and 39.11\% in deep urban canyons.
\end{abstract}

\section{Introduction}
\label{sec:introduction}

Localization is a fundamental requirement for autonomous ground vehicles, such as robotic delivery and patrolling platforms \cite{reid_localization_2019}. To meet this need, GNSS provides continuous and global positioning, navigation, and timing (PNT) services, serving as the backbone of outdoor navigation \cite{groves_principles_2013}. In particular, GNSS real-time kinematic (RTK) uses double-differenced (DD) carrier-phase measurements to achieve centimeter-level accuracy once the carrier-phase ambiguities are resolved \cite{wen_towards_2021}.
However, GNSS signals remain vulnerable to multipath effects, signal blockage, and hardware faults, especially in complex environments such as urban canyons and dense forests \cite{shen_advancing_2024,sunderhauf_towards_2012,zhang_principles_2023}. These disturbances introduce large measurement errors, commonly known as outliers, across pseudorange, carrier-phase, and Doppler observables \cite{li_review_2022}. In RTK positioning, pseudorange outliers directly corrupt the initial float ambiguity estimates and degrade ambiguity resolution, especially during initialization or after signal outages \cite{wen_towards_2021}. Furthermore, these unmitigated outliers propagate into GNSS-aided systems, such as loosely or tightly coupled GNSS/INS/odometer integrations, causing severe velocity, heading, and position drift \cite{li_high-precision_2022}.
Consequently, detecting and mitigating GNSS outliers is critical for reliable PNT performance. Over the past decades, extensive research has addressed this challenge, and existing approaches can be broadly categorized into GNSS-only and sensor-aided strategies.

\textbf{\textit{GNSS-only outlier mitigation is general but limited to geometric information statistics.}}
In this work, we first focus on the outlier mitigation for the generated measurements after signal processing. In this phase, GNSS-only approaches typically include pre-processing and robust state estimation. The pre-processing method relies on the physical models, aiming to flag outliers before the data are passed into the estimator. The fundamental idea is to recognize part of the measurements using consensus check on single or multiple kinds of GNSS measurements.
The effectiveness of this step strongly depends on the information source used to form the prediction. Traditional GNSS-only pre-processing methods exploit internal redundancy of the measurements, such as pseudo-range checks \cite{li_inter-frequency_2025}, carrier-phase combination tests \cite{xu_modified_2020}, or Doppler cross-validation \cite{icking_doppler_2020}. These methods are attractive because they do not require external sensors, but they often suffer from degraded performance in challenging environments with limited satellite visibility or highly correlated errors. More recently, advanced estimation methods and machine learning–based techniques have been introduced to enrich the prediction model. For instance, \cite{guo_new_2023} utilizes an autoregressive integrated moving average (ARIMA) model together with a multilayer perceptron (MLP) to generate pseudo-GNSS time series, thereby enabling the identification of outliers that deviate from learned patterns.
The robust estimation often exploits residual analysis or statistical consistency checks tied to a particular estimation framework. For example, \cite{zair_outlier_2016} employs a particle filter to detect pseudo-range and Doppler outliers by analyzing deviations in the measurement innovation sequence. Such methods can be accurate in some applications, but their performance is tightly related to the assumed system model and estimation algorithm, making them less flexible across different application scenarios.

\textbf{\textit{Sensor-aided outlier mitigation is feasible but relies on accurate reference.}}
In addition to GNSS-only approaches, sensor-aided methods, leveraging information from motion sensors or perception maps to assist outlier detection, have open a new window for outlier detection and mitigation in GNSS challenging environments. For example, a 3D map is first used in \cite{zhong_outlier_2022} to detect the NLOS measurements. However, the 3D map could cost much time and money to generate, making it to difficult to maintain. \cite{sun_new_2021} employs the innovation sequence from a tightly coupled GNSS/IMU filter to mitigate pseudo-range outliers, where the IMU serves as a short-term reference. \cite{wang_robust_2023} proposes a robust iterated cubature Kalman filter to minimize the impact of GNSS outliers under NLOS and multipath conditions in GNSS/INS fusion. While this strategy can be effective, it also introduces new challenges: IMU measurements are subject to intrinsic errors such as bias, scale factor, and noise, which vary between devices and can drift over time. If these parameters are not properly calibrated or estimated online, the predicted GNSS measurement will be inaccurate, which in turn may lead to false alarms or missed detections. 
As a common sensor on the ground vehicle, an odometer could provide stable velocity measurements during operation in environments. Similar to IMU, an odometer could provide high-frequency measurements to resume the vehicle motion in two dimensions \cite{li_high-precision_2022}. Nevertheless, it is still not employed to enhance GNSS measurement quality checking. 

In this work, we use calibrated wheel measurements to support GNSS pseudorange screening when a Doppler reference is unavailable or insufficient. The first stage compares pseudorange changes with integrated Doppler-derived range rates. When this test cannot be formed, relative motion provides another reference. We obtain this motion from a joint IMU-odometer prediction using past accepted information, without including the current GNSS observations. The IMU and wheel factors constrain the same poses, so the predicted three-dimensional attitude is not obtained from wheel yaw integration alone.

We then compare the measured and predicted temporal changes of double-differenced pseudoranges. This construction makes the initial position error enter through the change in satellite geometry, while the relative-motion error enters through the current differential line of sight. We analyze these two effects and project the prediction-graph covariance into the detection gate. Inconsistent ranges are downweighted before entering the GNSS-RTK/INS/odometer estimator. The method therefore extends the available screening information during GNSS recovery, provided that the motion prediction remains informative.

The contributions of this work are summarized below.
\begin{enumerate}
    \item \textbf{Odometer-aided pseudorange screening.} We complement an established Doppler consistency test with a temporal double-difference test driven by joint IMU-odometer prediction. Calibrated wheel measurements constrain the motion prior, allowing the second stage to operate without requiring a valid Doppler test or a new GNSS position solution at the current epoch.
    \item \textbf{Error analysis for detection and mitigation.} We derive the first-order test error from perturbed antenna poses and separate outlier change, anchor error, joint position prediction and antenna-transfer errors. This analysis relates the initial position and motion uncertainty to the detection threshold and uses the resulting variance in a common chi-square test. Under a trusted-reference assumption, a minimum-inflation constraint then determines the effective code variance for both stages.
    \item \textbf{Evaluation of complementary screening conditions.} We examine the uncertainty model using local-graph Monte Carlo tests and organize reference navigation experiments with isolated pseudorange injections, recorded degraded trajectories and controlled GNSS interruptions. Doppler-availability and IMU-only controls distinguish the roles of the two references. The retained navigation results precede the revised weighting rule and are identified separately from its analytical derivation.
\end{enumerate}

\begin{figure}[!t]
    \centering
    \includegraphics[width=\columnwidth]{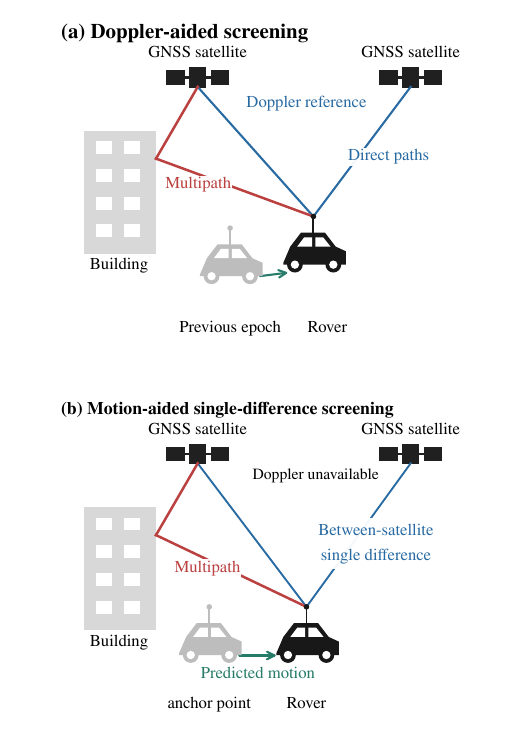}
    \caption{Two-stage pseudorange screening using Doppler and motion-aided between-satellite single differences. (a) Valid Doppler predicts the pseudorange change between consecutive epochs. (b) When Doppler is unavailable, joint IMU-odometer prediction provides the antenna displacement from the anchor point to the current rover position. Red paths illustrate reflected signals alongside the direct paths. In (b), the two satellites and the base and rover receivers form the double difference. The antenna increment transfers the accepted anchor to the current epoch. Paths are schematic and not to scale.}
    \label{fig:placeholder}
\end{figure}

\section{Preliminary Knowledge}
We retain the notation and calibrated motion model of \cite{song}. The frames $w,n,e,b,m$ denote world, east-north-up, ECEF, IMU and vehicle coordinates. The vehicle axes are right-forward-up. A rotation $\R_b^a$ maps vectors from $b$ to $a$, and $\boldsymbol\ell_{\mathrm{IG}}^b$ is the IMU-to-antenna lever arm (m). Epochs $a,k$ denote an accepted anchor and the current observation. Satellite indices are $i,j$, and all state subscripts denote discrete epochs. Physical time is $t_k$ and the interval is $\Delta t_k=t_k-t_{k-1}$ (s). RI denotes RTK/INS tightly coupled integration, and RIO adds the odometer integral factor. In our notation, hats represent measurements, and bars represent predicted states along with their propagated covariances. Superscripts denote only the corresponding reference frames. Similarly, the bound-derived residual variance carries a bar to separate it from the ideal variance $S_{\mathrm{mot}}$. To prevent self-validation in fault detection, current GNSS observations are excluded from the motion prediction.

\subsection{GNSS observation model}
The code and Doppler-derived range rate satisfy
\begin{equation}
\begin{aligned}
 \hat P_k^i&=\rho_k^i+c(\delta t_{r,k}-\delta t_k^i)+I_k^i+T_k^i+B_k^i+\epsilon_{P,k}^i,\\
 \dot P_{\mathrm{Dop},k}^i&=\dot\rho_k^i+c(\delta\dot t_{r,k}-\delta\dot t_k^i)
 -\dot I_k^i+\dot T_k^i+\epsilon_{\mathrm{Dop},k}^i,
\end{aligned}
\label{eq:gnssmodel}
\end{equation}
where the epoch index $k$ is suppressed in the following definitions. The quantity $\hat P^i$ is raw pseudorange (m), $\dot P_{\mathrm{Dop}}^i$ is Doppler-derived range rate (m/s), $r$ identifies the receiver, $\rho^i$ is geometric range (m), $c$ is light speed (m/s), and $\delta t_r,\delta t^i$ are receiver and satellite clock offsets (s). The terms $I^i,T^i,B^i,\epsilon_P^i$ denote ionospheric delay, tropospheric delay, code outlier error and nominal code noise (m). The dot denotes a rate sampled at epoch $k$, not a continuous-time state or a finite difference of the code, and $\epsilon_{\mathrm{Dop}}^i$ is range-rate error (m/s). Doppler follows carrier range rate, whose ionospheric term has the opposite sign to code \cite{groves_principles_2013}. In Stage 1, we compare the code increment against the integrated Doppler over one epoch, so their short-interval ionospheric divergence and integration error must remain small relative to the gate. Crucially, Doppler is not the derivative of the code outlier error $B^i$. This property makes Doppler a useful consistency reference for detecting code outliers, although it is not immune to tracking errors \cite{zair_outlier_2016,icking_doppler_2020,song_certifiable_2026}.

\subsection{Doppler-pseudorange consistency screening}
Between-satellite differencing removes the common receiver clock term from code-Doppler consistency tests \cite{yun_sdom_2022}. Applying this principle in Stage 1, we integrate the differenced Doppler range rate using the trapezoidal rule \cite{zhang_nlos_2024}, yielding
\begin{equation}
\begin{aligned}
 r_{\mathrm{Dop},k}^{ij}&=(\hat P_k^i-\hat P_k^j)-(\hat P_{k-1}^i-\hat P_{k-1}^j)\\
 &\quad-\frac{\Delta t}{2}\sum_{q\in\{k-1,k\}}(\dot P_{\mathrm{Dop},q}^i-\dot P_{\mathrm{Dop},q}^j),
\end{aligned}
\label{eq:dimplementation}
\end{equation}
where $r_{\mathrm{Dop},k}^{ij}$ is the consistency residual (m), $i$ and $j$ denote the tested and reference satellites, $q$ indexes the two endpoints, and $\Delta t=t_k-t_{k-1}$ is the sampling interval (s). To ensure hardware bias consistency, both satellites must belong to the same constellation and signal group $\mathcal G_k$. Furthermore, each receiver is tested separately so that the common receiver clock offset and drift terms cancel out through between-satellite differencing.

For independent endpoint and satellite errors, propagation through (\ref{eq:dimplementation}) gives
\begin{equation}
\begin{aligned}
 S_{\mathrm{Dop},i}&=\sigma_{P,k,i}^2+\sigma_{P,k-1,i}^2
 +\frac{\Delta t^2}{4}(\sigma_{\dot P,k,i}^2+\sigma_{\dot P,k-1,i}^2),\\
 S_{\mathrm{Dop}}^{ij}&=S_{\mathrm{Dop},i}+S_{\mathrm{Dop},j},
\end{aligned}
\label{eq:dvariance}
\end{equation}
where $S_{\mathrm{Dop},i}$ and $S_{\mathrm{Dop}}^{ij}$ denote the single-satellite increment variance and the pair-residual variance (m$^2$), respectively. The terms $\sigma_P$ and $\sigma_{\dot P}$ represent the standard deviations of the code (m) and the Doppler-derived range rate (m/s). Because the increment errors of different satellites are independent, the pair variance is obtained as the sum of the two single-satellite variances. In practice, a measured temporal code variance can replace the sum of the two epoch code variances when time correlation is non-negligible. Finally, note that consistency tests sharing the same reference satellite $j$ are statistically correlated.

Because differencing combines errors from two satellites, an inconsistent pair alone cannot identify the faulty satellite. To isolate the fault to satellite $i$, we assume that reference satellite $j$ and accepted historical endpoints contain no outliers, while retaining their nominal noise in (\ref{eq:dvariance}). If no credible reference exists, Stage 1 skips the test. Otherwise, the chi-square test below is applied under this noise model to identify faulty measurements.
Valid Doppler provides a precise short-term reference, but its availability depends on receiver tracking states. Specifically, acquisition provides only a coarse Doppler estimate, whereas frequency refinement and carrier tracking are required for a stable reference \cite{tang_doppler_2013,juang_tracking_2009}. Therefore, missing Doppler is not replaced by zero. This limitation motivates a second consistency reference from relative motion.

\subsection{IMU and odometer motion models}
The IMU measurements are modeled as
\begin{equation}
\begin{aligned}
 \hat{\boldsymbol\omega}_k^b&=\boldsymbol\omega_k^b+\bg{k}+\mathbf n_{g,k},
 \\
 \hat{\mathbf f}_k^b&=\mathbf f_k^b+\ba{k}+\mathbf n_{a,k},
\end{aligned}
\label{eq:imumodel}
\end{equation}
where the quantities are sampled at epoch $k$. With this index suppressed, $\boldsymbol\omega^b,\mathbf b_g,\mathbf n_g$ are angular rate, gyro bias and noise (rad/s). The quantities $\mathbf f^b,\mathbf b_a,\mathbf n_a$ are specific force, accelerometer bias and noise (m/s$^2$). Bias-corrected measurements propagate attitude and velocity, and velocity integration gives the IMU displacement.

For the right-forward-up vehicle frame, the calibrated wheel inputs are
\begin{equation}
 \mathbf v_{o,k}^m=\mathbf e_y(1+s_v)\hat v_{o,k},\qquad
 \boldsymbol\omega_{o,k}^m=\mathbf e_z(1+s_\omega)\hat\omega_{o,k},
\label{eq:wheelmodel}
\end{equation}
where $\mathbf e_y=[0,1,0]^T$ and $\mathbf e_z=[0,0,1]^T$ are unit vectors. The measured speed $\hat v_o$ is in m/s and yaw rate $\hat\omega_o$ is in rad/s. The dimensionless scale corrections $s_v,s_\omega$ follow \cite{song}, with zero denoting no correction. The mounting rotation and lever arm transfer these inputs to the IMU. Their calibrated wheel pre-integration yields a relative translation $\Delta\hat{\mathbf p}_{O,uv}^{b_u}$ (m) and quaternion $\Delta\hat q_{O,uv}$ between successive factor times $u,v$. These relative-pose measurements constrain the IMU poses in the prediction graph.

\section{Methodology}
We first predict GNSS antenna motion and its covariance from the joint INS-odometer estimate. We then compare the predicted and measured range changes and propagate the motion uncertainty through that comparison. The resulting variance determines the chi-square threshold used to detect and downweight inconsistent pseudoranges.

The subscripts $\mathrm I$ and $\mathrm G$ identify the IMU center and GNSS antenna, respectively. The quantities $\pI{k}$ and $\Rb{k}$ denote the IMU position (m) and IMU-to-world rotation, while $\pG{k}$ denotes the ECEF antenna position (m). Its predicted increment from $a$ to $k$ is $\bar{\mathbf d}^e$ (m). For each satellite pair, $y_k$ is the corrected code double difference (m), $h_k$ its geometric model (m), and $\mathbf g_k^e$ the differential LOS. Satellite-pair indices are omitted within a single test.

\subsection{Motion-based model}

Relative odometry has been used in GNSS consistency screening \cite{huang_lidar_2022}, and tight GNSS and IMU integration similarly leverages inertial propagation to bound state uncertainty during signal degradation \cite{li_tightly-coupled_2017}. We use calibrated IMU-odometer prediction to obtain the antenna positions and their uncertainty without including the GNSS observations being tested.
Current prediction graph uses the state blocks of \cite{song} within a sliding-window factor graph optimization (FGO) framework \cite{song_refgo_2026}. They contain IMU pose, velocity and biases, GNSS clock and ambiguity variables, and enabled odometer calibration parameters. Accepted history and current motion measurements give the predicted joint state $\bar{\mathbf X}$ and local covariance $\bar{\mathbf P}$. RI uses IMU factors for motion prediction, while RIO also includes calibrated odometer factors. The predicted GNSS antenna position is given by
\begin{equation}
 \bar{\mathbf p}_{\mathrm G,k}^{e}=\mathbf p_{\mathrm{ref}}^{e}
 +\mathbf C(\bar{\mathbf p}_{\mathrm I,k}^{w}+\bar{\mathbf R}_{b,k}^{w}\boldsymbol\ell_{\mathrm{IG}}^b),
\label{eq:antfactor}
\end{equation}
where $\mathbf p_{\mathrm{ref}}^{e}$ is the fixed ECEF position of the world-frame origin (m), and $\mathbf C=\R_n^e\R_w^n$ maps world vectors through local ENU to ECEF. The calibrated antenna lever arm is fixed. Position and attitude uncertainty therefore propagate through this same geometric relation.

For each prediction interval $s=(u,v)$, the required displacement and covariance are
\begin{equation}
\begin{aligned}
 \bar{\mathbf d}_s^e&=\bar{\mathbf p}_{\mathrm G,v}^{e}-\bar{\mathbf p}_{\mathrm G,u}^{e},\\
 \bar{\boldsymbol\Sigma}_{\Delta,s}&=(\mathbf H_v-\mathbf H_u)\bar{\mathbf P}
                       (\mathbf H_v-\mathbf H_u)^T,
\end{aligned}
\label{eq:actualcovariance}
\end{equation}
where $\bar{\boldsymbol\Sigma}_{\Delta,s}$ is the displacement covariance (m$^2$) and $\mathbf H_k=\partial\pG{k}/\partial\delta\mathbf X$ is its antenna-position Jacobian evaluated at the predicted state, with $\delta\mathbf X$ denoting the local perturbations of the joint state. For right attitude perturbations $\delta\thetaB{k}$ (rad), its nonzero blocks are $\mathbf H_{k,p}=\mathbf C$ and $\mathbf H_{k,\theta}=-\mathbf C\bar{\mathbf R}_{b,k}^{w}[\boldsymbol\ell_{\mathrm{IG}}^b]_\times$, where $[\mathbf z]_\times\mathbf a=\mathbf z\times\mathbf a$. The joint $\bar{\mathbf P}$ retains endpoint cross correlations and the effects of estimated biases and calibration. No separate bias-drift variance is added. The single-epoch antenna covariance is $\mathbf H_k\bar{\mathbf P}_k\mathbf H_k^T$.

Across successive prediction solves, a causal implementation retains increments without retrospectively smoothing the accepted anchor,
\begin{equation}
 \bar{\mathbf d}^e=\sum_{s\in\mathcal S_{ak}}\bar{\mathbf d}_s^e,
 \qquad\bar{\mathbf p}_{\mathrm G,k}^{e}=\bar{\mathbf p}_{\mathrm G,a}^{e}+\bar{\mathbf d}^e,
\label{eq:incrementchain}
\end{equation}
where $\mathcal S_{ak}$ is the gap-free interval chain from accepted epoch $a$ to current epoch $k$. An incomplete chain cannot support this test. A direct implementation instead subtracts the stored anchor position from the current graph prediction in (\ref{eq:antfactor}). Subsequent accepted updates can revise the direct prediction. The experiments specify which prediction is used.

\subsection{Uncertainty}
For a short baseline, common clock and atmospheric errors are assumed sufficiently removed by differencing. We compare the predicted geometric change with the measured code change, then propagate motion uncertainty through that comparison. Nominal receiver noise is retained without further modelling here.

Let $y_k=\nabla\Delta\hat P_k^{ij}$ be the rover-base, satellite-$i$-minus-$j$ code double difference (m), and $h_k(\mathbf p^e)=\nabla\Delta\|\mathbf p_{\mathrm S_i,k}^e-\mathbf p^e\|$ its geometric counterpart for a known base position. Here $\mathbf p_{\mathrm S_i,k}^e$ is satellite position and $\mathbf p^e$ is receiver position. The temporal double-difference residual is
\begin{equation}
\begin{aligned}
 r_{\mathrm{mot}}&=\left[y_k-h_k(\bar{\mathbf p}_{\mathrm G,k}^{e})\right]\\
 &\quad-\left[y_a-h_a(\bar{\mathbf p}_{\mathrm G,a}^{e})\right],
\end{aligned}
\label{eq:td}
\end{equation}
where $r_{\mathrm{mot}}$ is the triple-difference test residual (m).

To determine whether this discrepancy indicates an outlier, write the observation and predicted positions as
\begin{equation}
\begin{aligned}
 y_k&=h_k(\pG{k})+B_k+\nu_k,\\
 \bar{\mathbf p}_{\mathrm G,a}^{e}&=\pG{a}+\boldsymbol\epsilon_a^{e},\\
 \bar{\mathbf p}_{\mathrm G,k}^{e}&=\pG{k}+\boldsymbol\epsilon_a^{e}+\boldsymbol\epsilon_d^{e},
\end{aligned}
\label{eq:positionperturbed}
\end{equation}
where $B_k$ is the double-difference outlier, $\nu_k$ is nominal measurement and correction error, and $\boldsymbol\epsilon_a^{e},\boldsymbol\epsilon_d^{e}$ are anchor and antenna-displacement errors, all in m. The same anchor error enters both endpoints. Using $\partial h_k/\partial\mathbf p^e=-(\mathbf g_k^e)^T$ in (\ref{eq:td}) cancels the true geometric change and gives
\begin{equation}
\begin{aligned}
 r_{\mathrm{mot}}\simeq{}&\underbrace{\Delta B}_{\text{outlier change}}
 +\underbrace{(\Delta\mathbf g^e)^T\boldsymbol\epsilon_a^{e}}_{\text{anchor error}}\\
 &+\underbrace{(\mathbf g^e)^T\boldsymbol\epsilon_d^{e}}_{\text{joint motion error}}
 +\underbrace{\eta}_{\text{measurement error}},
\end{aligned}
\label{eq:error}
\end{equation}
where $\Delta B=B_k-B_a$ is the change in DD outlier and $\eta=\nu_k-\nu_a$ is the differenced nominal error, both in m. The unit vector $\mathbf u_k^{e,i}=(\mathbf p_{\mathrm S_i,k}^{e}-\pG{k})/\|\mathbf p_{\mathrm S_i,k}^{e}-\pG{k}\|$ points from rover to satellite $i$, $\mathbf g^e=\mathbf g_k^e=\mathbf u_k^{e,i}-\mathbf u_k^{e,j}$ is the dimensionless differential LOS, and $\Delta\mathbf g^e=\mathbf g_k^e-\mathbf g_a^e$. The anchor contribution is bounded by $\|\Delta\mathbf g^e\|\,\|\boldsymbol\epsilon_a^e\|$. A coarse position can support a short-interval test because its error is multiplied by LOS change. A long interruption increases this contribution as the satellite geometry changes. Constant outliers cancel between endpoints, and equal outlier changes on two satellites can also cancel. The test detects pair inconsistency rather than a unique faulty satellite.

Define $\boldsymbol\Sigma_d=\operatorname{Cov}(\boldsymbol\epsilon_d^{e})$ and $\boldsymbol\Sigma_{ad}=\operatorname{Cov}(\boldsymbol\epsilon_a^{e},\boldsymbol\epsilon_d^{e})$, both in m$^2$. Under zero-mean errors independent of the code error, (\ref{eq:error}) gives
\begin{equation}
\begin{aligned}
 S_{\mathrm{mot}}={}&(\Delta\mathbf g^e)^T\boldsymbol\Sigma_a\Delta\mathbf g^e
       +(\mathbf g^e)^T\boldsymbol\Sigma_d\mathbf g^e\\
       &+2(\Delta\mathbf g^e)^T\boldsymbol\Sigma_{ad}\mathbf g^e
       +\sigma_\eta^2,
\end{aligned}
\label{eq:variance}
\end{equation}
where $S_{\mathrm{mot}}$ is the ideal clean-residual variance (m$^2$), $\boldsymbol\Sigma_a$ is anchor-position covariance (m$^2$), and $\sigma_\eta^2$ is the variance of the differenced nominal code error (m$^2$). The cross term accounts for the common estimation history of anchor and motion. If code and prediction errors are correlated, their cross terms must also be included. The simulation below retains the full motion-chain covariance and uses independent anchor and code errors to check this expression.

Each interval covariance retains the endpoint cross correlations within its prediction graph. Across separate prediction solves, however, the stored blocks do not retain all increment-to-increment and anchor-to-increment cross correlations required by (\ref{eq:variance}). We therefore bound their combined effect by summing projected standard deviations,
\begin{equation}
\begin{aligned}
 \sigma_{\mathrm{mot}}&=\sum_{s\in\mathcal S_{ak}}
 \sqrt{(\mathbf g^e)^T\bar{\boldsymbol\Sigma}_{\Delta,s}\mathbf g^e}
 +\sqrt{(\Delta\mathbf g^e)^T\boldsymbol\Sigma_a\Delta\mathbf g^e},\\
 \bar S_{\mathrm{mot}}&=\sigma_{\mathrm{mot}}^2+\sigma_\eta^2,
\end{aligned}
\label{eq:actualgatevariance}
\end{equation}
where $\sigma_{\mathrm{mot}}$ is a conservative bound on motion-induced residual standard deviation (m). For the causal increment model with consistent marginal covariances and the same independent code-noise model, this construction gives $\bar S_{\mathrm{mot}}\geq S_{\mathrm{mot}}$. It allows the worst-case positive correlation rather than setting missing cross covariances to zero. For direct prediction, revisions not represented by these increment covariances require empirical consistency checks and are not covered by this bound. The resulting wider gate can reduce detection sensitivity. If the full cross correlations are retained, (\ref{eq:variance}) can be used directly. Nominal code variances can be supplied directly by the receiver-dependent SNR weighting model of goGPS \cite{realini_gogps_2013}. The triple-difference first forms a receiver-satellite double difference from four pseudoranges, then subtracts it between two epochs. It therefore contains $2\times2\times2=8$ raw code observations. Under independent raw code errors, their variances sum to $\sigma_\eta^2$, since each differencing coefficient is $+1$ or $-1$ and is squared in variance propagation. The experiments specify the receiver-dependent noise coefficients.

\subsection{Two-stage detection}
The Doppler and motion stages provide $(r_{\mathrm{Dop}},S_{\mathrm{Dop}})$ and $(r_{\mathrm{mot}},\bar S_{\mathrm{mot}})$, respectively. Both follow the same sequence. A valid reference gives the residual and its nominal variance, a chi-square test flags inconsistency, and a minimum-inflation rule assigns the code weight. For a zero-mean Gaussian residual $r$ with variance $S$, the test is
\begin{equation}
 \kappa^2=\chi^2_{1,1-\alpha},\qquad
 \tau^2=\kappa^2S,\qquad r^2>\tau^2,
\label{eq:gate}
\end{equation}
where $r$ and $\tau$ are in m, $S$ is in m$^2$, $\alpha$ is the nominal per-test false-alarm probability, and $\chi^2_{1,1-\alpha}$ is the one-degree-of-freedom quantile \cite{teunissen_quality_1990}. Under the null hypothesis, $r^2/S\sim\chi_1^2$. Thus the distribution supplies $\kappa^2$, while error propagation supplies $S$. With a valid bound $\bar S_{\mathrm{mot}}$, the statistic is generally scaled chi-square and the test is conservative under the Gaussian model. This per-test statement does not imply a joint false-alarm probability for several correlated tests or for references selected from the same residuals.

For an outlier change $b$, $r\sim\mathcal N(b,S)$. Its probability of exceeding the gate is
\begin{equation}
 \mathcal P_{\mathrm{det}}(b)=1-\Phi\!\left(\frac{\tau-b}{\sqrt S}\right)
 +\Phi\!\left(\frac{-\tau-b}{\sqrt S}\right),
\label{eq:pd}
\end{equation}
where $b$ is the change remaining after differencing (m), $\mathcal P_{\mathrm{det}}$ is detection probability, and $\Phi$ is the standard normal cumulative distribution function. With a conservative gate, $S$ in this probability remains the actual clean-residual variance. Reducing motion uncertainty improves detection at the same nominal false-alarm level.

For individual code weighting, we assume that the reference satellite, base observations and accepted historical endpoints contain no outliers. Their nominal errors remain in the residual variance. Doppler or motion-prediction uncertainty must also be adequately represented. Under these conditions, we assign the detected discrepancy to the current rover code. For receiver-local Doppler screening, the same construction applies to that receiver's current code.

Let $(r_i,S_i)$ denote either valid test for current code $i$, whose nominal variance is $\sigma_{P,i}^2$ (m$^2$). We keep all reference and prediction uncertainties unchanged and add independent effective uncertainty only to this code. Since its differencing coefficient is $+1$ or $-1$, variance propagation gives
\begin{equation}
 \widetilde S_i(\lambda_i)
 =S_i+\underbrace{(\lambda_i-1)\sigma_{P,i}^2}_{\text{additional current-code variance}},
\label{eq:weightvariance}
\end{equation}
where $\lambda_i\geq1$ is the dimensionless variance multiplier and $\widetilde S_i$ is the adjusted test variance (m$^2$). This is an effective covariance for weighting. A fixed outlier changes the mean and does not itself increase physical noise variance.

We choose the smallest inflation satisfying the consistency constraint,
\[
 \lambda_i^\star=\underset{\lambda\geq1}{\operatorname{arg\,min}}\ \lambda
 \quad\text{subject to}\quad
 \frac{r_i^2}{S_i+(\lambda-1)\sigma_{P,i}^2}\leq\kappa^2.
\]
The constraint gives $\lambda\geq1+(r_i^2-\kappa^2S_i)/(\kappa^2\sigma_{P,i}^2)$. Combining it with $\lambda\geq1$ yields the same closed form for both stages,
\begin{equation}
 \lambda_{\mathrm{mot},i}^\star
 =1+\frac{[(r_{\mathrm{mot}}^{ij})^2-\kappa^2\bar S_{\mathrm{mot}}^{ij}]_+}
                 {\kappa^2\sigma_{P,i}^2},
\label{eq:inflation}
\end{equation}
\begin{equation}
 \lambda_{\mathrm{Dop},i}^\star
 =1+\frac{[(r_{\mathrm{Dop}}^{ij})^2-\kappa^2S_{\mathrm{Dop}}^{ij}]_+}
                 {\kappa^2\sigma_{P,i}^2},
\label{eq:dweight}
\end{equation}
where $[x]_+=\max(x,0)$ follows from the nondecreasing-variance constraint, and $j$ is the trusted reference for the corresponding test. All residual variances use nominal code noise, before either stage changes the weight. The multiplier is one below the threshold. Above it, any smaller multiplier violates the constraint. No additional inflation limits are imposed. The chi-square model determines the boundary, while the stated optimization criterion determines the weight.

When both stages are valid for the same code, their constraints require
\begin{equation}
\begin{aligned}
 \lambda_i^\star&=\max(\lambda_{\mathrm{Dop},i}^\star,\lambda_{\mathrm{mot},i}^\star),\\
 \widetilde\sigma_{P,i}^2&=\lambda_i^\star\sigma_{P,i}^2,
 \qquad w_i=(\lambda_i^\star)^{-1},
\end{aligned}
\label{eq:combinedweight}
\end{equation}
where $\widetilde\sigma_{P,i}^2$ is effective raw-code variance (m$^2$) and $w_i$ is its relative information weight. The maximum is the smallest value satisfying both lower bounds. This does not assume independent tests. An unavailable test contributes no constraint and is represented by unit inflation. The binary flag $f_i$ is one if either valid test flags code $i$, and zero otherwise. It is retained after weighting. Double-difference covariance is rebuilt from the adjusted raw variances, including shared-reference terms. Carrier phase is screened separately.

Algorithm~\ref{alg:two-stage} applies the common test and minimum-inflation rule across the available references using chain prediction by default. The set $\mathcal D_k$ collects eligible temporal code/Doppler endpoints, excluding previously flagged measurements and unverified first samples after acquisition. The set $\mathcal V_k$ collects eligible tuples $(i,j,a)$ that possess a trusted reference satellite, a valid anchor $a$, stable IMU biases, and a complete chain of relative-motion increments. Before evaluating its residual, each candidate code selects one eligible reference per stage and its most recent valid anchor $a$ in Stage 2.

The anchor set $\mathcal A$ maintains accepted anchors updated exclusively by uninflated measurements and purges entries older than $T_{\max}$ (s). Because all detection tests evaluate residuals against nominal code variances, variance inflation in Stage 1 does not alter the Stage 2 detection gate.

\newcounter{algorithm}
\begin{figure}[!t]
\centering\textbf{Algorithm 1. Unified two-stage range screening}\par\smallskip\hrule\smallskip\refstepcounter{algorithm}
\label{alg:two-stage}
\small
\begin{algorithmic}[1]
\REQUIRE $\hat P,\dot P_{\mathrm{Dop}}$, IMU, odometer, calibration, $\alpha$.
\STATE $\mathcal A\gets\varnothing,\quad\kappa^2\gets\chi^2_{1,1-\alpha}$
\FOR{each epoch $k$}
 \STATE Predict $\bar{\mathbf X},\bar{\mathbf P}$ without current GNSS.
 \STATE $\bar{\mathbf d}_s^e,\bar{\boldsymbol\Sigma}_{\Delta,s}\gets(\ref{eq:actualcovariance})$
 \STATE $\bar{\mathbf d}^e\gets(\ref{eq:incrementchain})$ for chain prediction (or direct subtraction from (\ref{eq:antfactor})).
 \STATE $\lambda_{\mathrm{Dop},i}^\star\gets1,\quad\lambda_{\mathrm{mot},i}^\star\gets1\quad\forall i$
 \STATE $f_i\gets0\quad\forall i$
 \STATE \textit{Stage 1. Valid temporal Doppler reference.}
 \STATE Exclude flagged and unverified endpoints from $\mathcal D_k$.
 \FOR{$i\in\mathcal D_k$ with a trusted reference $j\ne i$}
  \STATE $(r_i,S_i)\gets(r_{\mathrm{Dop}}^{ij},S_{\mathrm{Dop}}^{ij})$, (\ref{eq:dimplementation}), (\ref{eq:dvariance})
  \STATE $f_i\gets1$ if $r_i^2>\kappa^2S_i$.
  \STATE $\lambda_{\mathrm{Dop},i}^\star\gets(\ref{eq:dweight})$
 \ENDFOR
 \STATE \textit{Stage 2. Valid motion and accepted anchor.}
 \FOR{$(i,j,a)\in\mathcal V_k$}
  \STATE $(r_i,S_i)\gets(r_{\mathrm{mot}}^{ij},\bar S_{\mathrm{mot}}^{ij})$, (\ref{eq:td}), (\ref{eq:actualgatevariance})
  \STATE $f_i\gets1$ if $r_i^2>\kappa^2S_i$.
  \STATE $\lambda_{\mathrm{mot},i}^\star\gets(\ref{eq:inflation})$
 \ENDFOR
 \STATE $\lambda_i^\star\gets\max(\lambda_{\mathrm{Dop},i}^\star,\lambda_{\mathrm{mot},i}^\star)$
 \STATE $\widetilde\sigma_{P,i}^2\gets\lambda_i^\star\sigma_{P,i}^2$
 \STATE Update navigation, preserving the outlier flags.
 \STATE Refresh $\mathcal A$ using accepted, noninflated ranges.
 \STATE $\mathcal A\gets\mathcal A\setminus\{a\mid t_k-t_a>T_{\max}\}$
 \STATE \textbf{output} $\{\hat P_i,\widetilde\sigma_{P,i}^2,f_i\}$
\ENDFOR
\end{algorithmic}\smallskip\hrule
\end{figure}

\section{Experimental Validation}
The experiments examine motion uncertainty, screening availability and positioning error. Dataset and method descriptions precede the simulation and real-test configurations.
\subsection{Experimental Setup}
\subsubsection{Datasets}
The Gugong ground-vehicle recording supplies raw rover and base GNSS observations, IMU measurements, 2D odometer data (wheel speed and yaw rate), and a postprocessed reference trajectory, with its platform and sensor calibration detailed in \cite{song}. In this recording, the open first loop supports controlled fault injections after convergence, whereas the later loop provides natural GNSS degradation without artificial faults.
UrbanNav-HK-Medium-Urban-1 and UrbanNav-HK-Deep-Urban-1 provide two further Hong Kong urban driving recordings \cite{hsu_urbannav_2023}, denoted UrbanNav Medium and UrbanNav Deep. We use their GNSS and IMU measurements along with the supplied postprocessed GNSS/inertial reference, evaluating only the RTK and RI methods due to the absence of wheel data.
Reference positions are transferred to the tested GNSS antenna using the attitude and lever arm, and positioning errors are evaluated directly at common epochs after initialization without trajectory alignment. Horizontal RMSE combines East and North errors, Up RMSE uses the vertical component, and 3D RMSE combines all three components. Estimator runtime is benchmarked on a single thread of a MacBook Air laptop with an M5 CPU (24 GB RAM).

\subsubsection{Compared methods}
RTK uses GNSS alone, RI adds the IMU, and RIO further includes calibrated wheel measurements. Each sensor configuration is evaluated without outlier screening and with Doppler-only screening, denoted OM1. RI and RIO are additionally evaluated with the motion-based stage, denoted OM2, isolating the individual contributions of Doppler screening, motion screening, and odometer aiding.
For external baselines, we adapt the outlier-detection schemes of Zair et al.\ \cite{zair_outlier_2016} and Sun et al.\ \cite{sun_new_2021} into our factor-graph backends rather than using their original filtering estimators. Specifically, we implement the multi-epoch kinematic fitting and a-contrario subset selection of Zair et al.\ \cite{zair_outlier_2016} on the RTK backend, and the IMU-aided innovation-window fault screening of Sun et al.\ \cite{sun_new_2021} on the RI backend to match their respective sensor inputs. Both external detectors adjust pseudorange weights through our common weighting interface, and all compared methods within each sensor group share the same carrier-phase observations, ambiguity handling, and nominal sensor parameters.

\subsubsection{Simulation configuration}
Two experiments examine different parts of the method. The first is a local Gaussian simulation of motion uncertainty and detection over 20,000 trials. An error model propagates along straight, level motion at 2 m/s with a 1-s interval over prediction durations of 1--60 s, and RIO includes odometer relative-pose constraints. Initial per-axis position, velocity, attitude, gyro-bias and accelerometer-bias standard deviations are 0.30 m, 0.03 m/s, $1.75\times10^{-3}$ rad, $1.75\times10^{-4}$ rad/s and 0.01 m/s$^2$, respectively. The antenna lever arm is $[0,0,0.47]^T$ m, and the triple-difference code noise standard deviation is 1.50 m. The initial satellite azimuth/elevation pairs are $(20^\circ,25^\circ)$ and $(210^\circ,70^\circ)$, and the differential LOS rotates about the Up axis at $1.50\times10^{-4}$ rad/s. Differenced outlier changes range from 0 to 25 m, and the full simulated covariance serves as the reference for uncertainty and detection checks.

The second experiment injects controlled faults into the first open Gugong loop using recorded motion, broadcast ephemerides, the initialized graph covariance and a maximum anchor age of 120 s. After a common convergence interval, thirty one-epoch code pulses are introduced at 10-s intervals, alternating between G10 and G25 on two signals with amplitudes of 10, 15 and 20 m. This yields 60 injected signal observations while leaving carrier phase unchanged. The three test conditions apply continuous GNSS, a 30-s outage or a 60-s outage immediately before the first pulse, where an outage removes GNSS observations while motion propagation continues. GPS and BeiDou support initialization, followed by GPS-only evaluation over the subsequent 300 s using 299 common epochs.

\begin{table*}[!tbp]
\centering\small
\caption{Method parameters for the two simulation experiments.}
\label{tab:simulationparameters}
\renewcommand{\arraystretch}{1.12}
\begin{tabular}{>{\raggedright\arraybackslash}p{0.14\textwidth}>{\raggedright\arraybackslash}p{0.29\textwidth}>{\raggedright\arraybackslash}p{0.24\textwidth}>{\raggedright\arraybackslash}p{0.23\textwidth}}
\toprule
Symbol & Meaning & Local Gaussian simulation & Gugong injection experiment\\
\midrule
$\alpha$ & Per-test false-alarm probability & $0.01$ & $0.05$\\
$\mathbf n_g$ & Gyro noise density (rad/s/$\sqrt{\mathrm{Hz}}$) & $3.49\times10^{-4}$ & $2.67\times10^{-4}$\\
$\mathbf n_a$ & Accelerometer noise density (m/s$^2$/$\sqrt{\mathrm{Hz}}$) & $0.03$ & $0.03$\\
$\Delta\hat{\mathbf p}_{O,uv}^{b_u}$ & Wheel translation SD (m) & $[0.03,0.03,0.20]^T$ & Propagated preintegration covariance\\
$\Delta\hat q_{O,uv}$ & Wheel rotation SD, local angles (rad) & $[1.75,1.75,0.87]^T\!\times10^{-2}$ & Propagated preintegration covariance\\
$\hat v_o$ & Wheel-speed noise density (m/s/$\sqrt{\mathrm{Hz}}$) & Included in relative-pose noise & $0.01$\\
$\hat\omega_o$ & Wheel yaw-rate noise density (rad/s/$\sqrt{\mathrm{Hz}}$) & Included in relative-pose noise & $0.10$\\
$\boldsymbol\Sigma_a$ & Anchor covariance (m$^2$) & $0.30^2\mathbf I_3$ & Accepted-graph covariance\\
\bottomrule
\end{tabular}
\par\smallskip\begin{minipage}{0.94\textwidth}\footnotesize
SD denotes standard deviation. Initial SDs specify the corresponding state-covariance blocks. Rotation SDs use local angle perturbations, not quaternion-component noise. The first simulation uses relative-pose noise directly, while the second propagates wheel input noise. Lever arms are expressed in the adopted $b$ frame of the estimator.
\end{minipage}
\end{table*}

\subsubsection{Real-test configuration}
The unmodified Gugong loop, UrbanNav Medium, and UrbanNav Deep contain 310, 597, and 1344 common epochs, respectively. Gugong provides IMU and 2D odometer data, whereas UrbanNav provides four-constellation GNSS (GPS, GLONASS, Galileo, and BeiDou) and IMU data without wheel measurements.

All tests use a 1-s GNSS interval and a 120-s maximum anchor age. In the body frame, the IMU-to-antenna lever arms are $[0.13,0.20,0.24]^T$ m for Gugong and $[0,-0.86,0.31]^T$ m for UrbanNav. Gyro and accelerometer noise densities are $2.67\times10^{-4}$ rad/s/$\sqrt{\mathrm{Hz}}$ and 0.03 m/s$^2$/$\sqrt{\mathrm{Hz}}$ for Gugong, and $1.03\times10^{-2}$ rad/s/$\sqrt{\mathrm{Hz}}$ and 0.01 m/s$^2$/$\sqrt{\mathrm{Hz}}$ for UrbanNav. Gugong uses wheel-speed and yaw-rate noise densities of 0.01 m/s/$\sqrt{\mathrm{Hz}}$ and 0.10 rad/s/$\sqrt{\mathrm{Hz}}$, along with the calibrated IMU-odometer extrinsic and scale parameters from \cite{song}. These Gugong settings also apply to its injection tests. For GNSS weighting, the Doppler range-rate standard deviation is 0.20 m/s, and nominal pseudorange variance follows the SNR-based goGPS model \cite{realini_gogps_2013}. The nominal false-alarm probability $\alpha$ is fixed at 0.01 for the Gaussian simulation, 0.05 for Gugong, and a stricter 0.001 for both UrbanNav sequences to suppress false alarms under severe urban multipath.

\subsection{Simulation Tests}
\subsubsection{Results}
\textit{Uncertainty and detection simulation.}
To isolate the motion-aided Stage 2 detector without Doppler screening, we first compare predicted motion uncertainty with the errors generated by the local Gaussian model. Over tested durations from 1 to 60 s, empirical and predicted motion standard deviations differ by less than 1\%, and the mean normalized squared residual stays within 0.99 to 1.01. With the exact residual variance, observed false-alarm rates are 0.89\% to 1.11\%, closely matching the prescribed 1\% level. Fig.~\ref{fig:covbound} shows the motion errors and their confidence intervals, where the joint-covariance 95\% intervals achieve mean pointwise coverages of 94.83\% for RI and 94.93\% for RIO.
\fig{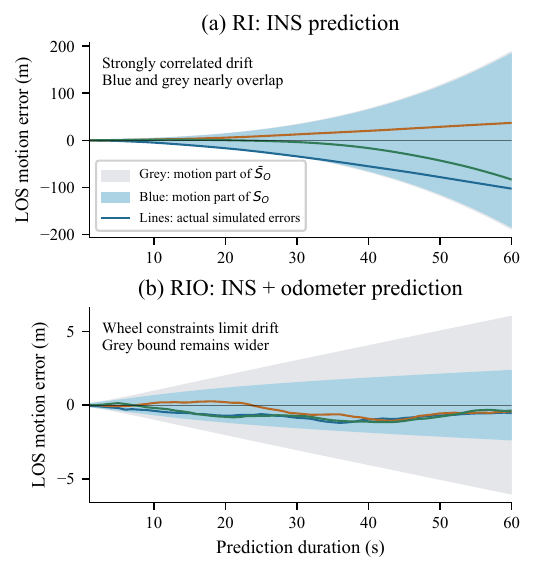}{LOS motion errors in three trials of the local Gaussian model. Blue bands show the joint-covariance 95\% interval and grey bands show the accumulated standard-deviation bound. Coverage is evaluated over 20,000 trials. Both intervals contain only the motion contributions to $S_{\mathrm{mot}}$ and $\bar S_{\mathrm{mot}}$, excluding anchor and code noise. The panels use different vertical scales.}{fig:covbound}
As the prediction duration increases, the LOS motion standard deviations of RI and RIO diverge sharply. At 30 and 60 s, RI reaches 16.09 and 93.65 m, whereas RIO stays at 0.76 and 1.18 m. For RIO, the accumulated standard-deviation bounds are 1.52 and 3.06 m, which yield total detection thresholds of 5.51 and 8.78 m after adding anchor and code uncertainty in (\ref{eq:gate}).
Fig.~\ref{fig:mcstage2} evaluates the resulting Stage 2 detection performance when Doppler aiding is absent. At 30 s, a 10-m outlier change is detected in only 2.08\% of RI trials, compared with 99.58\% of RIO trials. At 60 s, RIO still detects 73.72\% of 10-m changes and 99.95\% of 15-m changes. Its analytical detection probabilities from (\ref{eq:pd}) are 99.62\% and 73.83\% at 30 and 60 s, confirming our theoretical derivation.
\fig{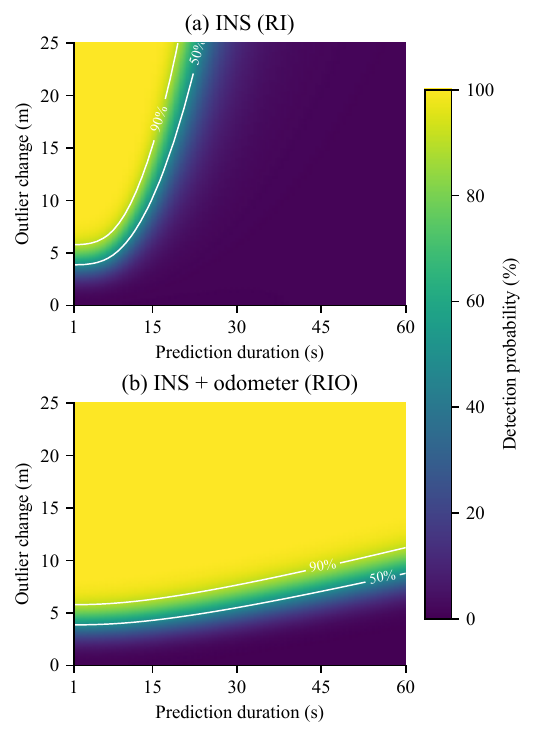}{Detection probability without Doppler in the local Gaussian model. RI and RIO share nominal false-alarm probability and measurement noise. White contours indicate 50\% and 90\% detection. Probabilities follow (\ref{eq:pd}) and are checked against 20,000 Gaussian trials. They do not depend on the code-weighting rule.}{fig:mcstage2}

\begin{figure*}[!tbp]\centering
\includegraphics[width=\textwidth]{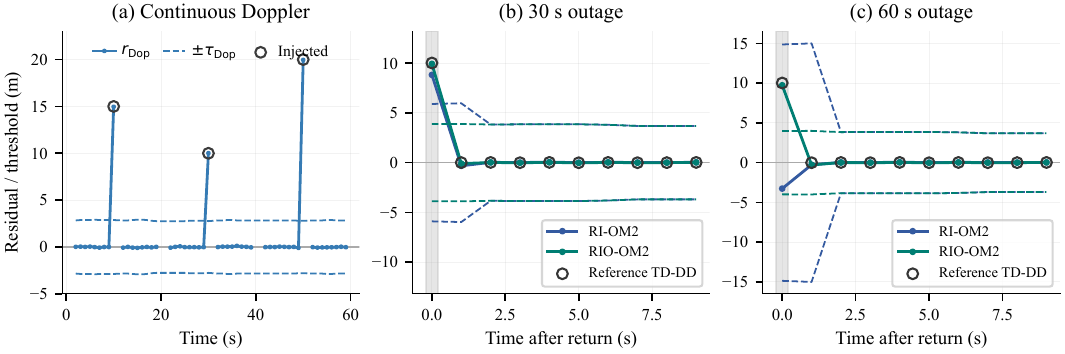}
\caption{Controlled code-pulse detection in the injection replays. (a) G25, 1C with continuous Doppler, where open circles mark injected outliers. (b), (c) G10:G25, 1C after 30-s and 60-s outages, where open circles show reference triple-difference from antenna truth and the actual detector anchor, solid curves show test residuals and dashed curves show gates. Grey marks the first injected epoch.}
\label{fig:currentdetection}\end{figure*}

\textit{Controlled injection experiment.}
For the injection replays, Table~\ref{tab:currenttiming} reports detection counts and estimator-update times. Under continuous GNSS tracking, Stage 1 Doppler screening (RTK-OM1) and Zair et al.\ \cite{zair_outlier_2016} both detect all 60 injected observations, leaving no missed faults for Stage 2 to resolve. After either outage, however, the first two returning signals lack a valid prior Doppler epoch, so RTK-OM1 and Zair et al.\ both flag 58 pulses and miss the two first-return faults. In contrast, RI-OM2 and RIO-OM2 rely on Stage 2 to detect all 60 pulses across all conditions, including both first-return faults. Sun et al.\ \cite{sun_new_2021} flags 58 pulses and captures both first-return faults after the 30-s outage, but drops to 52 flagged pulses and misses both first-return faults after the 60-s outage. As shown in Fig.~\ref{fig:currentdetection}(b), both RI-OM2 and RIO-OM2 detect the first returning pulse after the 30-s outage. After the 60-s outage, however, the RI-OM2 residual on G10:G25 falls inside its inflated gate, whereas the RIO-OM2 residual closely tracks the true triple difference and triggers detection. Note that the aggregate counts in Table~\ref{tab:currenttiming} cover all tested satellite pairs across the 60 injected observations, and therefore differ from the single-pair trace in Fig.~\ref{fig:currentdetection}.

\begin{table*}[!tbp]\centering\small
\caption{Detection counts and estimator-update time in the injection replays. Total counts use 60 injected observations, and first-return counts use the two injected signals at the first epoch after GNSS recovery. Timing uses the 30-s outage experiment over 299 common epochs, where P95 denotes the 95th percentile.}
\begin{tabular}{lcccccrr}\toprule
& \multicolumn{3}{c}{Total flagged / 60} & \multicolumn{2}{c}{First return / 2} & \multicolumn{2}{c}{Update time (ms)}\\
Method & 0 s & 30 s & 60 s & 30 s & 60 s & Mean & P95\\\midrule
RTK & 0 & 0 & 0 & 0 & 0 & 2.39 & 3.66\\
RTK-OM1 & 60 & 58 & 58 & 0 & 0 & 2.18 & 3.05\\
Zair et al.~\cite{zair_outlier_2016} & 60 & 58 & 58 & 0 & 0 & 2.94 & 4.23\\
\midrule
RI & 0 & 0 & 0 & 0 & 0 & 14.61 & 17.23\\
RI-OM1 & 60 & 58 & 58 & 0 & 0 & 17.33 & 26.43\\
RI-OM2 & 60 & 60 & 60 & 2 & 2 & 32.03 & 50.80\\
Sun et al.~\cite{sun_new_2021} & 58 & 58 & 52 & 2 & 0 & 20.39 & 24.23\\
\midrule
RIO & 0 & 0 & 0 & 0 & 0 & 36.15 & 41.40\\
RIO-OM1 & 60 & 58 & 58 & 0 & 0 & 44.34 & 67.84\\
RIO-OM2 & 60 & 60 & 60 & 2 & 2 & 79.36 & 126.47\\
\bottomrule\end{tabular}\label{tab:currenttiming}\end{table*}
Table~\ref{tab:currentsimulation} summarizes the resulting positioning errors. With continuous tracking, RTK-OM1 reduces 3D RMSE from \resZeroRTKThreeD{} to \resZeroRTKSThreeD{} m, outperforming Zair et al.\ by \resZeroZairGain\%. Because Stage 1 alone captures all pulses under continuous Doppler, OM1 and OM2 yield nearly identical accuracy for both RI (\resZeroRISThreeD{} versus \resZeroRIDThreeD{} m) and RIO (\resZeroRIOSThreeD{} versus \resZeroRIODThreeD{} m).
After the 30-s outage, where Stage 1 and Zair et al.\ miss the first returning pulses and yield 3D RMSEs of \resThirtyRTKSThreeD{} and \resThirtyZairThreeD{} m, Stage 2 enables RI-OM2 to reduce 3D RMSE from \resThirtyRIThreeD{} to \resThirtyRIDThreeD{} m and improve on Sun et al.\ by \resThirtySunGain\%, while RIO-OM2 reaches \resThirtyRIODThreeD{} m. After the 60-s outage, RI-OM2 achieves \resSixtyRIDThreeD{} m and improves on Sun et al.\ (\resSixtySunThreeD{} m) by \resSixtySunGain\%. Meanwhile, RIO-OM2 reduces 3D RMSE by \resSixtyRIOGain\% from its \resSixtyRIOThreeD{}-m baseline to \resSixtyRIODThreeD{} m, with Up RMSE dropping from \resSixtyRIOUp{} to \resSixtyRIODUp{} m, clearly outperforming Doppler-only RIO-OM1 (\resSixtyRIOSThreeD{} m).
In terms of computational efficiency in Table~\ref{tab:currenttiming}, RI-OM2 and RIO-OM2 require mean update times of \resTimeRIDMean{} and \resTimeRIODMean{} ms, with 95th percentiles of \resTimeRIDPninetyfive{} and \resTimeRIODPninetyfive{} ms. These runtimes stay well within the 1-s GNSS sampling interval, confirming real-time capability.

\begin{table*}[!tbp]\centering\small
\caption{Positioning errors in the code-pulse injection replays. Horizontal / Up / 3D RMSE (m) over 299 common epochs. Bold marks the minimum within each sensor group and component.}
\begin{tabular}{lccc}\toprule
Method & No outage & 30 s outage & 60 s outage\\
\midrule
RTK & 2.12 / 7.63 / 7.92 & 2.13 / 8.02 / 8.30 & 2.13 / 8.02 / 8.30\\
RTK-OM1 & \textbf{0.20} / \textbf{0.22} / \textbf{0.30} & \textbf{0.29} / \textbf{1.91} / \textbf{1.94} & \textbf{0.29} / \textbf{1.92} / \textbf{1.94}\\
Zair et al.~\cite{zair_outlier_2016} & 0.31 / 0.66 / 0.73 & 0.35 / 1.95 / 1.99 & 0.35 / 1.96 / 1.99\\
\midrule
RI & 0.18 / 0.63 / 0.66 & 0.50 / 2.09 / 2.15 & 0.42 / 2.79 / 2.82\\
RI-OM1 & \textbf{0.16} / \textbf{0.22} / \textbf{0.28} & 0.39 / 0.60 / 0.72 & 0.28 / 1.72 / 1.74\\
RI-OM2 & \textbf{0.16} / 0.23 / \textbf{0.28} & \textbf{0.23} / \textbf{0.30} / \textbf{0.38} & \textbf{0.21} / \textbf{0.34} / \textbf{0.40}\\
Sun et al.~\cite{sun_new_2021} & 0.19 / 0.25 / 0.31 & \textbf{0.23} / 0.45 / 0.51 & 1.15 / 0.69 / 1.34\\
\midrule
RIO & \textbf{0.11} / 0.74 / 0.75 & 0.22 / 1.00 / 1.02 & 0.28 / 1.24 / 1.27\\
RIO-OM1 & \textbf{0.11} / \textbf{0.18} / \textbf{0.21} & 0.18 / 0.19 / 0.26 & 0.23 / 0.23 / 0.33\\
RIO-OM2 & \textbf{0.11} / \textbf{0.18} / \textbf{0.21} & \textbf{0.17} / \textbf{0.15} / \textbf{0.23} & \textbf{0.21} / \textbf{0.17} / \textbf{0.27}\\
\bottomrule\end{tabular}\label{tab:currentsimulation}\end{table*}

\begin{figure*}[!tbp]\centering
\includegraphics[width=\textwidth,height=0.84\textheight,keepaspectratio]{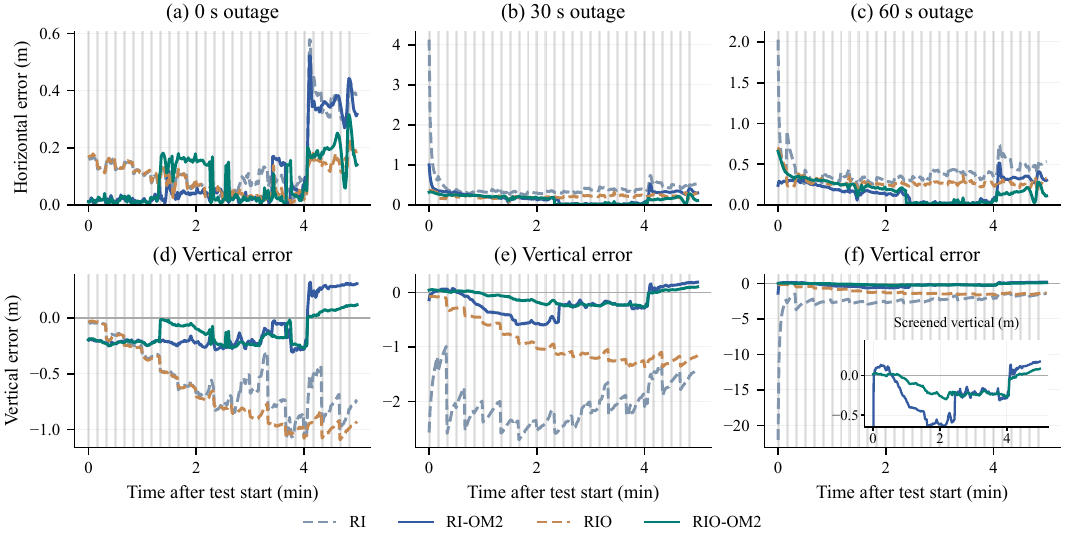}
\caption{Horizontal and signed vertical errors in the injection replays. Grey marks identify injected epochs. The inset enlarges screened vertical errors after 60 s. Each column uses the common epochs of its condition in Table~\ref{tab:currentsimulation}.}
\label{fig:currentsimulation}\end{figure*}

\subsubsection{Discussion}
\textit{Uncertainty and detection simulation.}
The close match between empirical and predicted errors validates our covariance propagation under linear Gaussian assumptions. Although RI and RIO show similar 95\% coverage, their interval widths differ by two orders of magnitude because persistent attitude and bias errors strongly correlate RI increments over time. Adding wheel constraints suppresses this drift in RIO, while the accumulated standard-deviation bound conservatively covers worst-case positive correlations across solves. Consequently, RIO's false-alarm rate at 30 s drops from 1.02\% under the exact variance to 0.12\% under the bound, trading a slightly higher detectable outlier threshold at 60 s (Fig.~\ref{fig:mcstage2}) for guaranteed conservatism.

\textit{Controlled injection experiment.}
The injection results clarify the complementary roles of the two screening stages and the advantages over external baselines. When Doppler tracking is continuous, Stage 1 alone captures every injected pulse, outperforming Zair et al.\ through tighter minimum-inflation weighting rather than aggressive subset rejection. Right after a GNSS outage, newly reacquired signals lack valid Doppler history, which causes both Stage 1 and the GNSS-only screening of Zair et al.\ to miss the first returning faults. By leveraging inertial motion prediction, Stage 2 immediately screens outliers upon signal recovery. Compared with the innovation-window screening of Sun et al., which loses sensitivity as inertial errors accumulate over the 60-s outage, our propagated motion uncertainty and conservative bound preserve full fault detection across all tested pairs. Furthermore, the divergence between RI-OM2 and RIO-OM2 on individual pairs after the 60-s outage confirms that wheel aiding keeps the residual gate tight during extended interruptions. Finally, as shown in Fig.~\ref{fig:currentsimulation}, positioning accuracy depends not only on detection counts but also on how inflated code weights balance against the motion prior during state updates.

\subsection{Real Tests}
\subsubsection{Results}
\begin{figure*}[!tbp]\centering
\includegraphics[width=\textwidth]{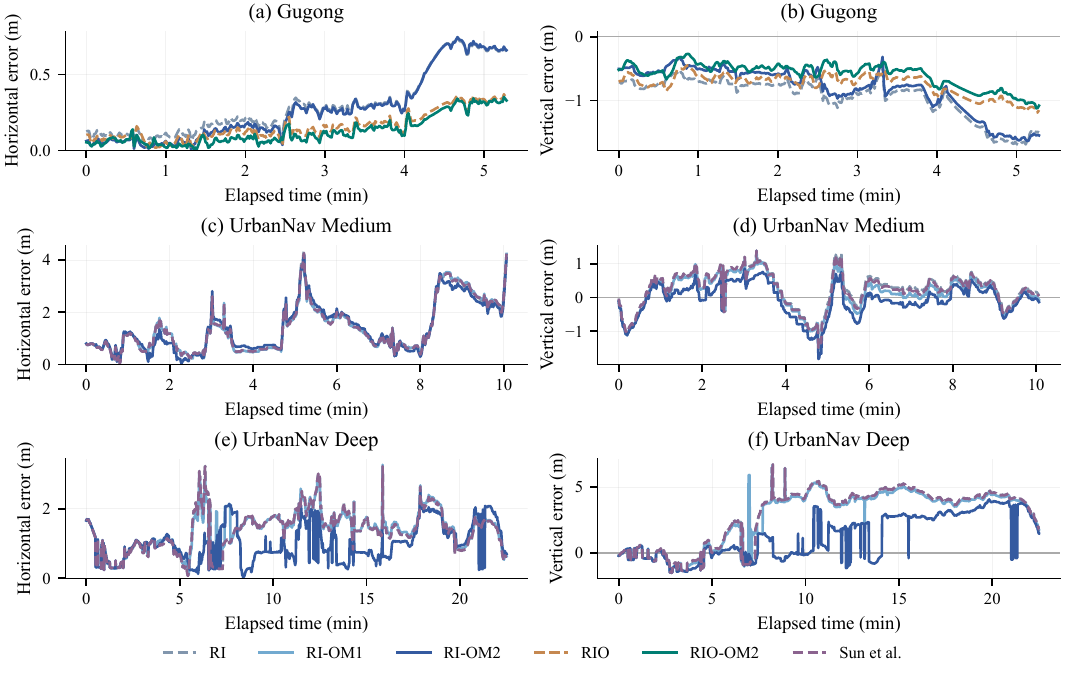}
\caption{Horizontal and signed vertical errors on unmodified field recordings. Rows cover the common supports in Table~\ref{tab:currentrecorded}. No trajectory alignment or mean-error removal is applied.}
\label{fig:currentrecorded}\end{figure*}

Table~\ref{tab:currentrecorded} summarizes the positioning accuracy on the three unmodified field sequences, and Fig.~\ref{fig:currentrecorded} shows the corresponding error time series. Doppler-only screening (OM1) slightly improves 3D RMSE on UrbanNav Medium and Deep, while leaving the Gugong baseline unchanged because gradual multipath there rarely triggers single-epoch Doppler checks. In contrast, adding the motion-based stage (OM2) consistently reduces positioning errors across all three recordings.

On Gugong, RIO-OM2 reduces horizontal RMSE from \resGugongRIOH{} to \resGugongRIODH{} m, Up RMSE from \resGugongRIOUp{} to \resGugongRIODUp{} m, and 3D RMSE from \resGugongRIOThreeD{} to \resGugongRIODThreeD{} m, achieving a \resGugongRIOGain\% overall reduction. While RI-OM2 (\resGugongRIDThreeD{} m) performs comparably to Sun et al.\ (\resGugongSunThreeD{} m), RIO-OM2 achieves the lowest error across all components by leveraging wheel constraints to bound motion uncertainty during natural GNSS degradation.

On UrbanNav Deep, severe urban canyon multipath and frequent signal re-acquisitions degrade Doppler continuity. Under these conditions, RI-OM2 reduces Up RMSE from \resDeepRIUp{} to \resDeepRIDUp{} m and 3D RMSE from \resDeepRIThreeD{} to \resDeepRIDThreeD{} m, outperforming Sun et al.\ by \resDeepSunGain\%. On UrbanNav Medium, where open-sky visibility is better and baseline errors are already low, RI-OM2 still reduces 3D RMSE from \resMediumRIThreeD{} to \resMediumRIDThreeD{} m.

\begin{table*}[!tbp]\centering\small
\caption{Positioning errors on unmodified field recordings. Horizontal / Up / 3D RMSE (m) on common epochs. Bold marks the minimum within each sensor group and component. N/A denotes unavailable odometer data.}
\begin{tabular}{lccc}\toprule
Method & Gugong & UrbanNav Medium & UrbanNav Deep\\
\midrule
RTK & \textbf{1.98} / \textbf{5.76} / \textbf{6.09} & 2.68 / 5.12 / 5.78 & \textbf{3.08} / 7.98 / 8.55\\
RTK-OM1 & \textbf{1.98} / \textbf{5.76} / \textbf{6.09} & 2.64 / \textbf{4.87} / \textbf{5.54} & \textbf{3.08} / \textbf{7.83} / \textbf{8.41}\\
Zair et al.~\cite{zair_outlier_2016} & \textbf{1.98} / 6.12 / 6.43 & \textbf{2.62} / 4.96 / 5.61 & \textbf{3.08} / 7.98 / 8.55\\
\midrule
RI & 0.36 / 1.00 / 1.06 & 1.65 / 0.57 / 1.75 & 1.48 / 3.61 / 3.90\\
RI-OM1 & 0.36 / 1.00 / 1.06 & 1.65 / 0.53 / 1.74 & 1.50 / 3.47 / 3.78\\
RI-OM2 & \textbf{0.35} / \textbf{0.89} / 0.96 & \textbf{1.63} / \textbf{0.48} / \textbf{1.70} & \textbf{1.13} / \textbf{2.08} / \textbf{2.37}\\
Sun et al.~\cite{sun_new_2021} & \textbf{0.35} / \textbf{0.89} / \textbf{0.95} & 1.65 / 0.56 / 1.75 & 1.48 / 3.60 / 3.89\\
\midrule
RIO & 0.18 / 0.76 / 0.78 & N/A & N/A\\
RIO-OM1 & 0.18 / 0.76 / 0.78 & N/A & N/A\\
RIO-OM2 & \textbf{0.16} / \textbf{0.62} / \textbf{0.64} & N/A & N/A\\
\bottomrule\end{tabular}\label{tab:currentrecorded}\end{table*}

\subsubsection{Discussion}
The field tests confirm that the mechanisms observed in simulation transfer directly to complex urban environments. First, improvements appear most prominently in the vertical component, because weak vertical satellite geometry amplifies the impact of multipath-induced pseudorange outliers on altitude estimation. Second, the contrast between UrbanNav Medium and UrbanNav Deep highlights the value of two-stage screening. In open or semi-urban streets, Doppler measurements remain continuous and Stage 1 handles most isolated faults. In deep urban canyons, frequent building blockages repeatedly reset carrier tracking and disable temporal Doppler checks, allowing Stage 2 to screen outliers upon signal re-acquisition.

Similarly, the Gugong results demonstrate why sensor-aided motion quality governs Stage 2 effectiveness in practice. Under sustained foliage and building occlusion, IMU-only prediction in RI-OM2 gradually widens its residual gate, whereas wheel odometry in RIO-OM2 preserves tight relative motion bounds and sustains high detection sensitivity. Finally, unlike controlled pulse injections, real-world datasets contain unmodeled non-Gaussian errors without ground-truth fault labels. As indicated by (\ref{eq:error}), a constant bias that persists unchanged between the anchor and current epochs cancels out in temporal differencing. Maintaining a clean anchor set $\mathcal A$ and credible motion covariance is therefore essential to prevent error accumulation, even though no consistency test can guarantee complete detection of time-invariant biases.

\section{Conclusion}
This paper presents a two-stage GNSS pseudorange outlier mitigation method for tightly coupled GNSS-RTK/INS/odometer navigation. Stage 1 exploits Doppler measurements as a short-term range-change reference, whereas Stage 2 uses joint IMU-odometer motion prediction to screen newly reacquired signals when temporal Doppler checks are unavailable. By separating anchor, relative-motion, and measurement uncertainties, the detector propagates prediction covariance directly into the test gate and applies a unified minimum-inflation weighting rule across both stages. Experimental results across simulation and field datasets confirm the effectiveness of the proposed framework. 

Currently, the proposed consistency tests primarily target step-like outliers and rely on Gaussian uncertainty assumptions, leaving persistent or slowly varying biases undetectable when they cancel across temporal endpoints. Future work will integrate learning-based models to characterize complex non-Gaussian urban multipath distributions and directly estimate or repair pseudorange residuals under consecutive outlier conditions.

\bibliographystyle{IEEEtran}
\bibliography{references}

\end{document}